\documentclass{article}
\usepackage{authblk}
\usepackage{amsthm, amsmath, amsfonts, amssymb}
\usepackage{graphicx} 
\usepackage{multirow}
\usepackage{hyperref}

\DeclareMathAlphabet{\pazocal}{OMS}{zplm}{m}{n}
\DeclareMathOperator*{\argmin}{arg\,min}

\def\Rset{\ensuremath{\mathbb R}} 
\def\Nset{\ensuremath{\mathbb N}} 

\newcommand\Xset{\ensuremath{\mathcal X}} 
\newcommand\Yset{\ensuremath{\mathcal Y}} 
\newcommand\Gset{\ensuremath{\mathcal G}} 

\newcommand\A{\ensuremath{\mathcal A}} 

\newcommand\Q{\ensuremath{\mathbb Q}} 
\newcommand\E{\ensuremath{\mathbb E}} 
\newcommand\V{\ensuremath{\mathbb V}} 

\newcommand\loss{\ensuremath{\mathcal L}} 
\newcommand\pred{\ensuremath{\mathcal P}} 
\newcommand\stab{\ensuremath{\mathcal S}} 
\newcommand\simp{\ensuremath{\mathbb S}} 

\newcommand\I{\ensuremath{\mathcal I}} 
\newcommand\ind[1]{\mathbf{1}_{#1}}
\providecommand{\keywords}[1]{\textbf{\textit{Keywords: }} #1}

\theoremstyle{plain}
\newtheorem{definition}{Definition}[section]

\theoremstyle{definition}

\theoremstyle{remark}

\title{A New Method to Compare the Interpretability of Rule-based Algorithms}
\author{Vincent Margot$^{1}$ and George Luta$^{2}$}
\affil{%
$^{1}$  Advestis, 69 Boulevard Haussmann, F-75008 Paris, France; vmargot@advestis.com\\
$^{2}$  Department of Biostatistics, Bioinformatics and Biomathematics, Georgetown University, Washington, DC 20057-1484, USA; george.luta@georgetown.edu}

\date{}

\begin{document}
	\maketitle
	\begin{abstract}
	Interpretability is becoming increasingly important for predictive model analysis. Unfortunately, as remarked by many authors, there is still no consensus regarding this notion. The goal of this paper is to propose the definition of a score that allows to quickly compare interpretable algorithms. This definition consists of three terms, each one being quantitatively measured with a simple formula: \emph{predictivity}, \emph{stability} and \emph{simplicity}. While predictivity has been extensively studied to measure the accuracy of predictive algorithms, stability is based on the Dice-Sorensen index for comparing two rule sets generated by an algorithm using two independent samples. The simplicity is based on the sum of the lengths of the rules derived from the predictive model. The proposed score is a weighted sum of the three terms mentioned above. We use this score to compare the interpretability of a set of rule-based algorithms and tree-based algorithms for the regression case and for the classification case.
\end{abstract}


\section{Introduction}

The widespread use of machine learning (ML) methods in many important areas such as health care, justice, defense or asset management has underscored the importance of interpretability for the decision-making process. In recent years, the number of publications on interpretability has increased exponentially. For a complete overview of interpretability in ML the reader may see the book \cite{Molnar20} and the article \cite{molnar20interpretable}. We distinguish two main approaches to generate interpretable prediction models.
	
	The first approach is to use a non-interpretable ML algorithm to generate the predictive model, and then create a so-called \emph{post-hoc} interpretable model. One common solution is to use graphic tools, such as the Partial Dependence Plot (PDP) \cite{Friedman01} or the Individual Conditional Expectation (ICE) \cite{goldstein15}. A drawback of  these methods is that they are limited by the human perception. Indeed, a plot with more than 3 dimensions cannot be interpreted by humans and so it is not useful for data sets with many features. An alternative way consists in using a surrogate model to explain the model generated by a black-box. We refer to the algorithms Local Interpretable Model-agnostic Explanations (LIME) \cite{Ribeiro16}, DeepLIFT \cite{shrikumar17} and SHapley Additive exPlanations (SHAP) \cite{Lundberg17} that attempt to measure the importance of a feature in the prediction process (we refer to \cite{Guidotti18} for an overview of the available methods). However, as outlined in \cite{Rudin19}, the explanations generated by these algorithms may not be sufficient to allow a reasonable decision process.

	The second approach is to use {intrinsically} interpretable algorithms to directly generate interpretable models. There are two main families of intrinsically interpretable algorithms: tree-based algorithms that are based on decision trees such as Classification And Regression Trees (CART) \cite{CART}, Iterative Dichotomiser 3 (ID3) \cite{Quinlan86}, C4.5 \cite{Quinlan93}, M5P \cite{wang96}, Logistic Model Trees (LMT) \cite{landwehr05} and rule-based algorithms that are generating rule sets such as Repeated Incremental Pruning to Produce Error Reduction (RIPPER) \cite{Cohen95}, First Order Regression (FORS) \cite{Karalivc97}, M5 Rules \cite{Holmes99}, RuleFit \cite{Friedman08}, Ensemble of Decision Rules (Ender) \cite{Dembczynski08}, Node Harvest \cite{Meinshausen10} or more recently Stable and Interpretable RUle Set (SIRUS) \cite{Benard21,Benard21_2} and the Coverage Algorithm \cite{Margot21}. It is important to note that any tree can be converted into a set of rules, while the opposite is not true.

	These algorithms generate predictive models based on the notion of a \emph{rule}. A rule is an \emph{If-Then} statement of the form:
	\begin{eqnarray*}
	\text{IF} && c_1 \text{ And } c_2 \text{ And } \dots \text{ And } c_k\\
	\text{THEN} && \text{Prediction} = p, \nonumber
	\end{eqnarray*}
	The condition part \emph{If} is a logical conjunction, where the $c_i$'s are tests that check whether the observation has the specified properties or not. The number $k$ is called the \emph{length} of the rule. If all $c_i$'s are fulfilled the rule is said to be \emph{activated}. The conclusion part \emph{Then} is the prediction of the rule if it is activated.
	
	Even though rule-based algorithms and tree-based algorithms seem to be easy to understand, there is no exact mathematical definition for the concept of interpretability. This is due to the fact that interpretability involves multiple concepts as explained in \cite{Lipton18}, \cite{Doshi17}, \cite{Yu20} and \cite{Murdoch19}. The goal of this paper is to propose a definition that combines these concepts in order to generate an interpretability score. It is important to note that related concepts such as \emph{justice}, \emph{ethics}, and \emph{morality}, which are associated with specific applications to health care, justice, defense or asset management, cannot be measured quantitatively.

	As proposed in \cite{Yu20} and \cite{Benard21}, we describe an interpretability score for any model formed by rules based on the triptych \emph{predictivity}, \emph{stability}, and \emph{simplicity}:
The predictivity score measures the accuracy of the generated prediction model. The accuracy ensures a high degree of confidence in the generated model.
The stability score quantifies the sensitivity of an algorithm to noise, and it allows to evaluate the \emph{robustness} of the algorithm.
The simplicity score could be conceptualized as the ability to easily verify the prediction. A simple model makes it easy to evaluate some qualitative criteria such as \emph{justice}, \emph{ethics} and \emph{morality}. By measuring these three concepts we are therefore able to evaluate the interpretability of several algorithms for a given problem.

A similar idea has been proposed in \cite{hammer04} in the area of the Logical Analysis of Data (LAD), by introducing the concept of Pareto-optimal patterns or strong patterns. The main part of the LAD is to select the best patterns from the dataset based on a triptych that includes simplicity, selectivity, and evidence. The authors identify two extreme cases of patterns: strong prime patterns and strong spanned patterns. The first are the most specific strong patterns while the last are the simplest strong patterns. In \cite{Alexe08}, the authors have studied the effects of pattern filtering on classification accuracy. They show that the prime patterns do provide somewhat higher classification accuracy, although the loss of accuracy by using strong spanned patterns is relatively small. For an overview of the LAD we refer the readers to \cite{Alexe07}.

\section{Predictivity score}
The aim of a predictive model is to predict the value of a random variable of interest $Y \in \Yset$, given features $X \in \Xset$ where $\Xset$ is a $d$-dimensional space. Formally, we consider the standard setting as follows: Let $(X,Y)$ be a random vector in $\Xset \times \Yset$ of unknown distribution $\Q$ such that
	\begin{equation*}
	Y = g^*(X) + Z,
	\end{equation*}
	where $\E[Z] = 0$ and $\V(Z) = \sigma^2$ and $g^*$ is a measurable function from $\Xset$ to $\Yset$.
	
	We denote by $\Gset$ the set of all measurable functions from $\Xset$ to $\Rset$. The accuracy of a predictor $g \in \Gset$ is measured by its risk, defined as
	\begin{equation}\label{eq:risk}
	\loss(g) := \E_\Q \left[ \gamma\left(g; (X, Y) \right)\right],
	\end{equation}
	where $\gamma : \Gset \times (\Xset \times \Yset) \to [0, \infty [$ is called a contrast function and its choice depends on the nature of $Y$. The risk measures the average discrepancy between $g(X)$ and $Y$, given a new observation $(X, Y)$ from the distribution $\Q$. As mentioned in \cite{Arlot10}, the definition \eqref{eq:risk} includes most cases of the classical statistical models.
	
	Given a sample $D_n = \left( (X_1, Y_1), \dots, (X_n, Y_n) \right)$, our aim is to predict $Y$ given $X$. The observations $(X_i, Y_i)$ are assumed to be independent and identically distributed (i.i.d) from the distribution $\Q$.
	
We consider a statistical algorithm which is a measurable mapping from $(\Xset \times \Yset)^n$ to a class of measurable functions $\Gset_n \subseteq \Gset$. This algorithm generates a predictor $g_n$ by using the Empirical Risk Minimization principle (ERM) \cite{Vapnikbook}, meaning that
\begin{equation*}
	g_n = \argmin_{g \in \Gset_n} \loss_n(g),
\end{equation*}
where $\loss_n(g) = \frac1n\sum_{i=1}^{n} \gamma(g, \left(X_i, Y_i)\right)$ is the empirical risk.

	The notion of predictivity is based on the ability of an algorithm to provide an accurate predictor. This notion has been extensively studied before. In this paper we define the predictivity score as
\begin{equation}\label{eq:pred}
	\pred_n(g_n, h_n) := 1 - \dfrac{\loss_n(g_n)}{\loss_n(h_n )},
\end{equation}
where $h_n$ is a baseline predictor chosen by the analyst. The idea is to consider a na\"ive and easily built predictor chosen according to the contrast function.

For instance, if $Y \in \Rset$, we generally use the quadratic contrast with $\gamma\left(g; (X, Y) \right) = \left( g(X) - Y \right)^2$. In this case, the minimizer of the risk \eqref{eq:risk} is the regression function defined by
$$g^*(X) = \E_\Q \left[ Y \mid X \right], \text{ hence we set } h_n = \dfrac1n \sum_{i=1}^n Y_i.$$

If $Y \in \left\{0, 1\right\}$, we use the $0-1$ contrast function $\gamma\left(g; (X, Y) \right) := \ind{g(X) \neq Y}$, and the minimizer of the risk is the Bayes classifier defined by
$$g^*(X) = \ind{\Q(Y=1 \mid X) \geq 1/2}, \text{ hence we set } h_n =  \ind{\sum_{i=1}^n Y_i \geq n/2}.$$

	The predictivity score \eqref{eq:pred} is a measure of accuracy which is independent of the range of $Y$. The risk \eqref{eq:risk} is a positive function, so $\pred_n(g_n, h_n) < 1$. Moreover, if $\pred_n(g_n, h_n) < 0$, it means that the predictor $g_n$ is less accurate than the chosen baseline predictor $h_n$. Thus, in this case it is better to use the predictor $h_n$ instead of $g_n$. Hence, we can assume that the predictivity score is a positive number between $0$ and $1$.

\section{q-Stability score}
Usually, stability refers to the stability of the prediction \cite{Vapnikbook}. Indeed, it has been shown that stability and predictive accuracy are closely connected (see for example \cite{bousquet02,poggio04}). In this paper we are more interested in the stability of the generated model. The importance of the stability for interpretability has been presented in \cite{Yu13}. Nevertheless, generating a stable set of rules is challenging as explained in \cite{Letham15}. In \cite{Benard21,Benard21_2}, the authors have proposed a measure of the stability for rule-based algorithms based on the following definition:
	\begin{quotation}
		\begin{itshape}
			``A rule learning algorithm is stable if two independent estimations based on two independent samples, drawn from the same distribution $\Q$, result in two similar lists of rules.''
		\end{itshape}
	\end{quotation}

	The $q$-stability score is based on the same definition. This concept is problematic for algorithms that do not use feature discretization and work with real values. Indeed, if the feature is continuous, the probability that a decision tree algorithm will cut on the same exact value for the same rule for two independent samples is zero. For this reason, this definition of stability is too stringent in this case. One way to avoid this problem is to discretize all continuous features. The discretization of features is a common solution to control the complexity of a rule generator. In \cite{Fayyad93}, for example, the authors use entropy minimization heuristics to discretize features and for the algorithms Bayesian rule lists (BRL) \cite{Letham15}, SIRUS \cite{Benard21,Benard21_2} and Rule Induction Partioning Estimator (RIPE) \cite{Margot18} the authors have discretized the features by using their empirical quantiles. We refer to \cite{Dougherty95} for an overview of the common discretization methods.
\newline

	In this paper, to generate the $q$-stability score, we consider a discretization process on the conditions of the selected rules based on the empirical $q$-quantile of the implied continuous features. Because this process is only used for the calculation of the $q$-stability, it does not affect the accuracy of the generated model.
	
	First, we discretize the continuous features that are involved in the selected rules. Let $q \in \Nset$ be the number of quantiles considered for the $q$-stability score and let $X$ be a continuous feature. An integer $p \in \{1, \dots, q\}$, called bin, is assigned to each interval $[x_{(p-1)/q}, x_{p/q}]$, where $x_{p/q}$ is the $p$-th $q$-quantile of $X$. A discrete version of the $X$ feature, designated by $Q_q(X)$, is constructed by replacing each value with its corresponding bin. In other words, a value $p_a$ is assigned to all $a \in X$ such that $a \in [x_{(p_a-1)/q}, x_{p_a/q}]$.

	Then, we extend this discretization to selected rules by replacing the interval boundaries of the individual tests $c_i$ with the corresponding bins. For example, the test $X \in [a, b]$ becomes $Q_q(X) \in [ p_a, p_b ]$, where $p_a$ and $p_b$ are such that $a \in [x_{(p_a-1)/q}, x_{p_a/q}]$ and $b \in [x_{(p_b-1)/q}, x_{p_b/q}]$.
	
	Finally, the formula for the $q$-stability score is based on the so-called Dice-Sorensen index. Let $\A$ be an algorithm and let $D_n$ and $D_n'$ be two independent samples of $n$ i.i.d. observations drawn from the same distribution $\Q$. We denote by $R_n$ and $R_n'$ the rule sets generated by an algorithm $\A$ based on $D_n$ and $D_n'$, respectively. Then, the $q$-stability score is calculated as
	\begin{equation}\label{eq:stab}
	\stab^q_n(\A) := \dfrac{2\left|Q_q(R_n) \cap Q_q(R_n') \right|}{|Q_q(R_n)| + |Q_q(R_n')|},
	\end{equation}
	where $Q_q(R)$ is the discretized version of the rule set $R$, with the convention that $0/0 = 0$, and the discretization process is performed by using $D_n$ and $D_n'$, respectively.
	
	The $q$-stability score \eqref{eq:stab} is the ratio of the common rules between $Q_q(R_n)$ and $Q_q(R_n')$. It is a positive number between $0$ and $1$: If $Q_q(R_n)$ and $Q_q(R_n')$ have no common rules, then $\stab^q_n(\A) = 0$, while if $Q_q(R_n)$ and $Q_q(R_n')$ have the same rules, then $\stab^q_n(\A) = 1$.

\section{Simplicity score}
Simplicity as a component of interpretability has been studied in \cite{luvstrek16} for the classification trees and in \cite{furnkranz20} for the rule-based models. In \cite{Margot21} the authors have introduced the concept of an \emph{interpretability index}, which is based on the sum of the length of all the rules of the prediction model. Such an interpretability index should not be confused with the broader concept of interpretability that is developed in this paper. As discussed in section \ref{sec:Inter}, the former will be interpreted as one of the components of the latter.
	\begin{definition}\label{def:inter}
		The interpretability index of an estimator $g_n$ generated by a rule set $R_n$ is defined by
		\begin{equation}\label{eq:inter_2}
		Int(g_n) := \sum_{r \in R_n} \text{length}(r).
		\end{equation}
	\end{definition}
	Even if \eqref{eq:inter_2} seems naive, we consider it to be a reasonable measure for the simplicity of a tree-based algorithm or a rule-based algorithm. Indeed, as the number of rules or the length of the rules increases, $Int(g_n)$ also increases. The fewer the number of rules and their lengths, the easier their understanding should be. 
	

	It is important to note that the value \eqref{eq:inter_2}, which is a positive number, cannot be directly compared to the scores from \eqref{eq:pred} and \eqref{eq:stab}, which are between $0$ and $1$.

	The simplicity score is based on the Definition \ref{def:inter}. The idea is to compare \eqref{eq:inter_2} relatively to a set of algorithms $\A_1^m = \{\A_1, \dots, \A_m\}$. Hence the simplicity of an algorithm $\A_i \in \A_1^m$ is defined in relative terms as follows:
	\begin{eqnarray}\label{eq:simp}
		\simp_n(\A_i, \A_1^m) = \dfrac{min\{Int(g_n^A: A \in \A_1^m)\}}{Int(g_n^{\A_i})}.
	\end{eqnarray}

	Similar to the previously defined scores, this quantity is also a positive number between $0$ and $1$: If $\A_i$ generates the simplest predictor among the set of algorithms $\A_1^m$ then $\simp_n(\A_i, \A_1^m) = 1$, and the simplicity of other algorithms in $\A_1^m$ are evaluated relatively to $\A_i$.

	We note that it would be useful to be able to calculate the simplicity score of only one algorithm. To do this, we would need to have a threshold value for the simplicity score. In practice this information could be obtained by using a survey on the maximum size of a rule set that people are willing to accept to use.
	
\section{Interpretability score}\label{sec:Inter}
	In \cite{Doshi17} the authors define interpretability as
	\begin{quotation}
		\begin{itshape}
	``the ability to explain or present to a person in an understandable form''
		\end{itshape}
	\end{quotation}
		  
		We claim that an algorithm with a high predictivity score \eqref{eq:pred}, stability score \eqref{eq:stab} and simplicity score \eqref{eq:simp} is interpretable in the sense of \cite{Doshi17}. Indeed, a high predictivity score ensures confidence in and truthfulness of the generated model, a high stability score ensures robustness and a good noise insensibility, and a high simplicity score ensures that the generated model is easy to understand for humans and can also be easily audited.

	The main idea behind the proposed definition of interpretability is to use a weighted sum of these three scores. Let $\A_1^m$ be a set of algorithms. Then, the interpretability of any algorithm $\A_i \in \A_1^m$ is defined as:
	\begin{equation}\label{eq:interpretability}
		\I(\A_i, D_n, D_n', \gamma, q) = \alpha_1 \pred(g_n^{\A_i}, \gamma) + \alpha_2 \stab^q_n(\A_i) + \alpha_3 \simp_n(\A_i, \A_1^m),
	\end{equation}
	where the coefficients $\alpha_1, \alpha_2$ and $\alpha_3$ have been chosen according to the analyst's objective, such that $\alpha_1 + \alpha_2 + \alpha_3 = 1$.

	It is important to note that the definition of interpretability \eqref{eq:interpretability} depends on the set of algorithms under consideration and the specific setting. Therefore, the interpretability score only makes sense within this set of algorithms and for the given setting.

\section{Application}
	The goal of this application is to compare several algorithms which are considered interpretable: Regression Tree \cite{CART}, RuleFit (RF) \cite{Friedman08}, NodeHarvest (NH) \cite{Meinshausen10}, Covering Algorithm (CA) \cite{Margot21} and SIRUS \cite{Benard21_2} for regression settings, and RIPPER \cite{Cohen95}, PART \cite{Frank98} and Classification Tree \cite{CART} for classification settings\footnote{We have excluded algorithms developed only for binary classification, such as M5Rules \cite{RWeka}, NodeHarvest \cite{Meinshausen10}, and SIRUS \cite{Benard21}}.
	
	\subsection{Brief overview of the selected algorithms}	
	RIPPER is a sequential coverage algorithm. It is based on the "divide-and-conquer" approach. This means that for a selected class it searches for the best rule according to a criterion and removes the points covered by that rule. Then it searches for the best rule for the remaining points and so on until all points of this class are covered. Then it moves on to the next class, with the classes being examined in the order of increasing size.

	PART is also a "divide-and-conquer" rule learner. The main difference is that in order to create the "best rule", the algorithm uses a pruned decision tree and keeps the leaf with the largest coverage.

	RuleFit is a very accurate rule-based algorithm. First it generates a list of rules by considering all nodes and leaves of a boosted tree ensemble ISLE \cite{Friedman03}. Then the rules are used as additional binary features in a sparse linear regression model that is using the Lasso \cite{Tibshirani96}. A feature generated by a rule is equal to $1$ if the rule is activated, and it is $0$ otherwise.

	NodeHarvest also uses a tree ensemble as a rule generator. The algorithm considers all nodes and leaves of a Random Forest as rules and solves a linear quadratic problem to fit a weight for each node. Hence, the estimator is a convex combination of the nodes.

	Covering Algorithm has been designed to generate a very simple model. The algorithm extracts a sparse rule set considering all nodes and leaves of a tree ensembles (using the Random Forest algorithm, Gradient Boosting algorithm \cite{Friedman01} or Stochastic Gradient Boosting algorithm \cite{Friedman02}). Rules are selected according to their statistical properties to form a "quasi-covering". The covering is then turned into a partition using the so-called \emph{partitioning trick} \cite{Margot18} to form a consistent estimator of the regression function.

	SIRUS has been designed to be a stable predictive algorithm. SIRUS uses a modified Random Forest to generate a large number of rules, and selects rules with a redundancy greater than the tuning parameter $p_0$. To be sure that redundancy is achieved, the features are discretized.

	For a comprehensive review of rule-based algorithms we refer to \cite{RuleLearningFundations,Furnkranz15}, while for a comprehensive review of interpretable machine learning we refer to \cite{Molnar20}.

\subsection{Datasets}
	We have used publicly available databases from the UCI Machine Learning Repository \cite{Dua2019} and from \cite{ElementST}. We have selected six datasets for regression\footnote{For the dataset \emph{Student} we have removed variables $G1$ and $G2$ which are the first and the second grade, respectively, because the target attribute $G3$ has a strong correlation with the attributes $G2$ and $G1$. In \cite{cortez08} the authors specify that it is more difficult to predict $G3$ without $G2$ and $G1$, although such prediction is much more useful.} which are summarized in Table \ref{tab:reg_datasets}, and three datasets for classification which are summarized in Table \ref{tab:classif_datasets}.
	\begin{table}
	\centering
	 \caption{\label{tab:reg_datasets} Presentations of the publicly available regression datasets used in this paper}
	\begin{tabular}{lcp{7.5cm}}
		\hline
		Name & $(n \times d)$ & Description \\
		\hline
		Ozone & $330 \times 9 $ & Prediction of atmospheric ozone concentration from daily meteorological measurements \cite{ElementST}. \\[0.2cm]
		Machine & $209 \times 8$ & Prediction of published relative performance \cite{Dua2019}. \\[0.2cm]
		MPG & $398 \times 8$ & Prediction of city-cycle fuel consumption in miles per gallon \cite{Dua2019}. \\[0.2cm]
		Boston & $506 \times 13$ & Prediction of the median price of neighborhoods, \cite{Harrison78}. \\[0.2cm]
		Student & $649 \times 32$ & Prediction of the final grade of the student based on attributes collected by reports and questionnaires \cite{cortez08}.\\[0.2cm]
		Abalone & $4177 \times 7$ & Prediction of the age of abalone from physical measurements \cite{Dua2019}.\\
		\hline
 	\end{tabular}
 	\end{table}

 	\begin{table}
 	\centering
 	\caption{\label{tab:classif_datasets} Presentations of the publicly available classification datasets used in this paper}
	\begin{tabular}{lcp{7cm}}
		\hline
		Name & $(n \times d)$ & Description \\
		\hline
		Wine &$4898 \times 11$ & Classification of white wine quality from $0$ to $10$ \cite{Dua2019}.\\[0.2cm]
		Covertype &$581012 \times 54 $&Classification of forest cover type $[1,7]$ based on cartographic variables \cite{Dua2019}.\\[0.2cm]
		Speaker &$329 \times 12$ &Classification of accent, six possibilities, based on features extracted from the first reading of a word \cite{Fokoue20}.\\
		\hline
 	\end{tabular}
 	\end{table}

\subsection{Execution} 	
 	For each dataset we perform $10$-fold cross-validation. The parameter settings for the algorithm are summarized in Table \ref{tab:algo_params}. These parameters were selected according to the author's recommendations to generate models based on rules with equivalent lengths. It means that all rules generated by algorithms have a bounded length. The parameters of the algorithms are not tuned because it is not the purpose of the paper to rank the algorithms. The aim of this section is to illustrate how this score is computed.
 	
 	For each algorithm, a model is fitted on the training set to obtain the simplicity score \eqref{eq:inter_2}, while we measure the predictivity score \eqref{eq:pred} on the test set. To obtain the predictivity score we set
 	\begin{align*}
 	&\gamma\left(g; (X, Y) \right) = \left( g(X) - Y \right)^2 &\text{and}& &&h_n = \dfrac1n \sum_{i=1}^n y_i  &&\text{for regression},\\
 	&\gamma\left(g; (X, Y) \right) = \ind{g(X) \neq Y} &\text{and}& &&h_n = mode\left( \{y_1, \dots, y_n\} \right) &&\text{for classification}.
 	\end{align*}
 	
 	Then, to obtain the stability score, the training set is randomly divided into two sets of equal length and two models are constructed. The code is a combination of Python and R and it is available on GitHub \href{https://github.com/Advestis/Interpretability}{https://github.com/Advestis/Interpretability}.

	\begin{table}
	\centering
	\caption{\label{tab:algo_params} Algorithms parameter settings}
	\begin{tabular}{lp{7.1cm}}
		\hline
		Algorithm & Parameters \\
		\hline
		CART & $max\_leaf\_nodes = 20$.\\[0.2cm]
		RuleFit & $tree\_size=4$, \newline $max\_rules=2000 $. \\[0.2cm]
		NodeHarvest & $max.inter=3$. \\[0.2cm]
		CA & $generator\_func=RandomForestRegressor$,\newline $n\_estimators = 500$,\newline $max\_leaf\_nodes=4$,\newline $alpha=1/2-1/100$, \newline$gamma=0.95$, \newline $k\_max = 3$\\[0.2cm]
		SIRUS & $max.depth=3$, \newline $num.rule = 10$. \\
		\hline
 	\end{tabular}
 	\end{table}
	
	The choice of $\alpha$'s in \eqref{eq:interpretability} is an important step in the process of comparing interpretability. For these applications we use an equally weighted average. It means that
	\begin{align*}
 	&\alpha_1 = 1/3, \qquad \alpha_2 = 1/3, \qquad \alpha_3 = 1/3.
 	\end{align*}
 	
 	Another possibility is to set each $\alpha$ to be inversely proportional to the variance of the associated score for each data set. In our application the results were very similar to the equally weighted case (data not shown).
 	 

\subsection{Results for regression}
\begin{figure}
\includegraphics[scale=0.45]{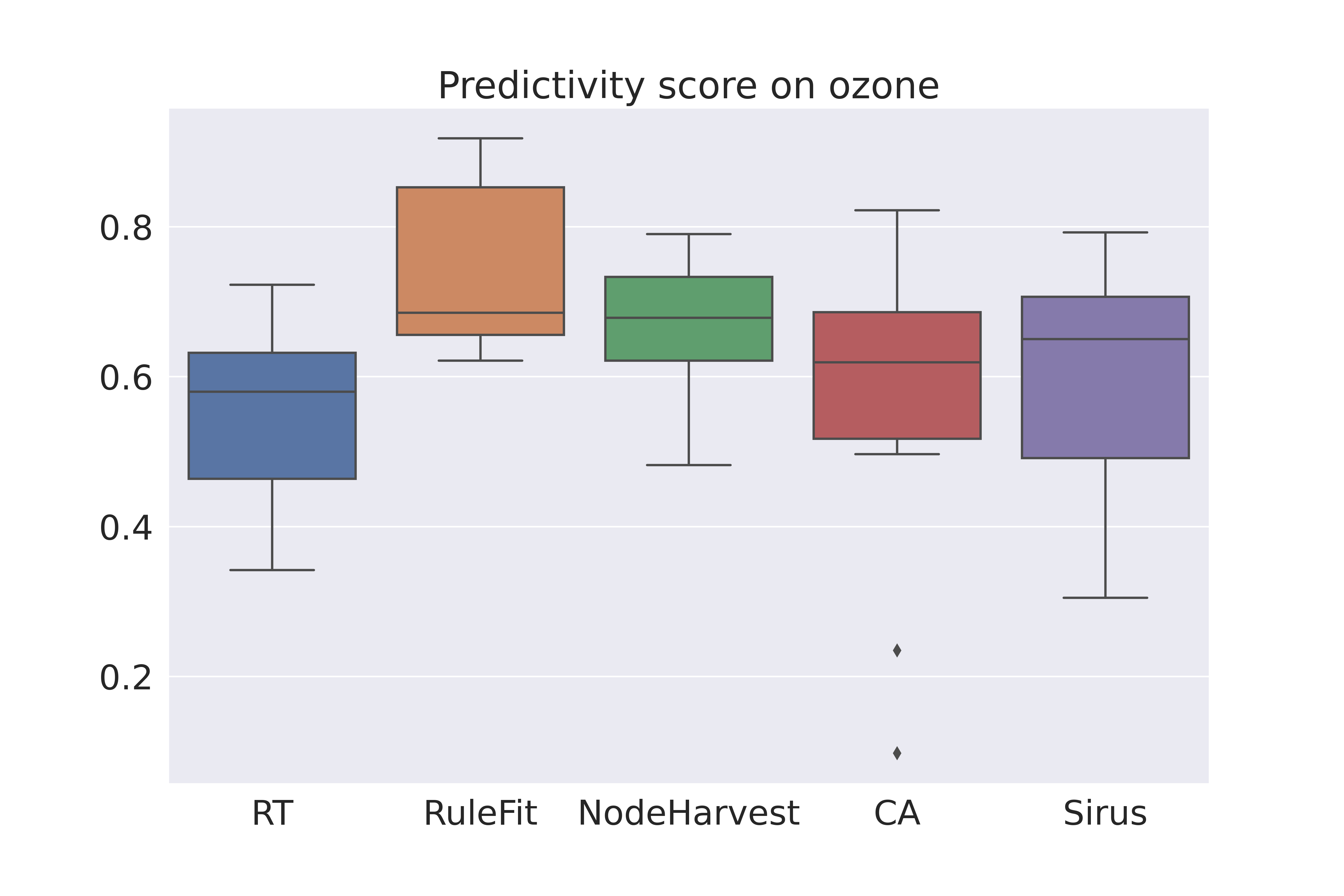}
\caption{\label{fig:pred} Box-plot of the prediction scores for each algorithm for the dataset Ozone}
\end{figure}

\begin{figure}

\includegraphics[scale=0.45]{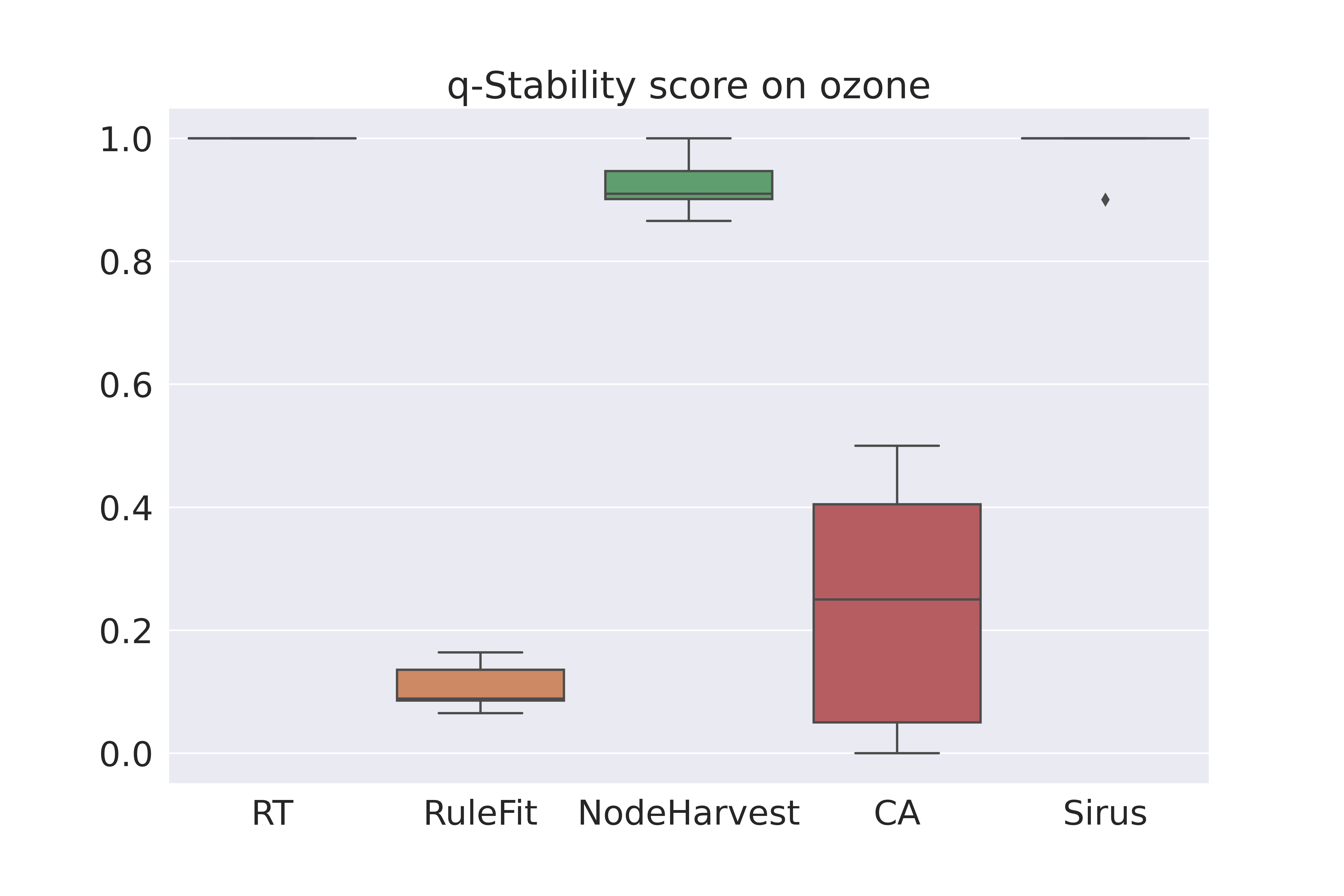}
\caption{\label{fig:stab} Box-plot of the $q$-stability scores for each algorithm for the dataset Ozone}
\end{figure}

\begin{figure}
\includegraphics[scale=0.45]{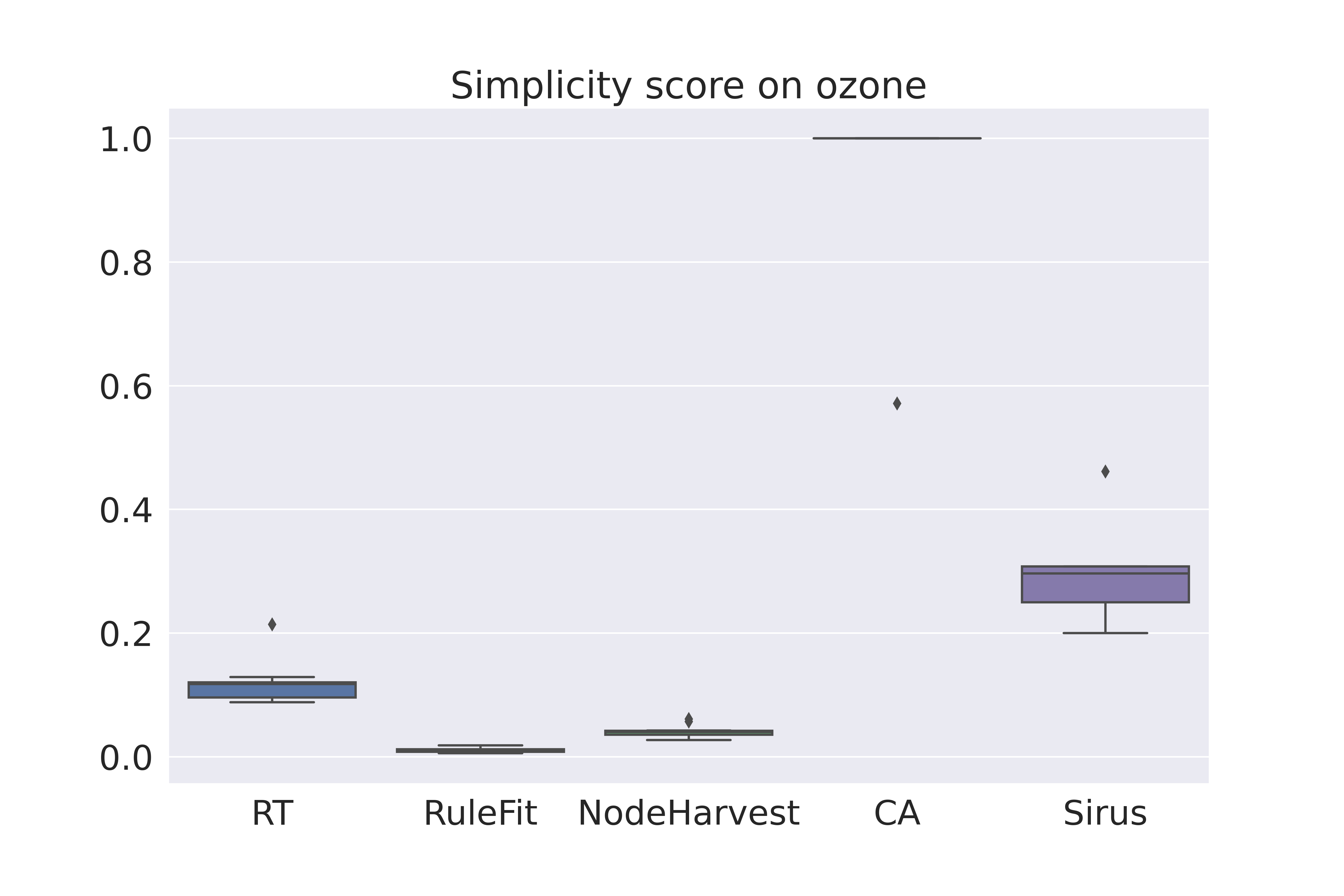}
\caption{\label{fig:simp} Box-plot of the simplicity scores for each algorithms for the data set Ozone}
\end{figure}

 	The averaged scores are summarized in Table \ref{tab:res_reg}. As expected RuleFit is the most accurate algorithm. However, RuleFit is neither stable nor simple. SIRUS is the most stable algorithm and the Covering Algorithm is one of the simplest. For all datasets, SIRUS seems to be the most interesting algorithm among this selection of algorithms and by our score \eqref{eq:interpretability}. Figures \ref{fig:pred}, \ref{fig:stab} and \ref{fig:simp} are the box-plots of the predictivity scores, $q$-stability scores and simplicity scores, respectively, of each algorithms on the dataset Ozone.

\begin{table}
\centering
\caption{\label{tab:res_reg} Average of predictivity score ($\pred_n$), stability score ($\stab_n^q$), simplicity score ($\stab_n$) and interpretability score ($\I$) over a $10$-fold cross-validation of commonly used interpretable algorithms for various public regression datasets. Best values are in bold, as well as values within 10\% of the maximum value for each dataset.}
		\begin{tabular}{lccccc}
			\hline
			\multirow{2}{*}{Dataset} & \multicolumn{5}{c}{$\pred_n$ } \\
			 & RT & RuleFit & NodeHarvest & CA & SIRUS \\
			\hline
			Ozone & 0.55 & \textbf{0.74} & 0.66 & 0.56 & 0.6 \\
			Machine & 0.79 & \textbf{0.95} & 0.73 & 0.59 & 0.46 \\
			MPG & 0.75 & \textbf{0.85} & \textbf{0.78} & 0.59 & 0.74 \\
			Boston & 0.61 & \textbf{0.74} & \textbf{0.67} & 0.26 & 0.57 \\
			Student & 0.08 & 0.16 & \textbf{0.22} & 0.13 & \textbf{0.24}  \\
			Abalone & 0.4 & \textbf{0.55} & 0.37 & 0.39 & 0.3 \\
			\hline
		\end{tabular}
		\vspace*{0.3cm}
		
		\begin{tabular}{lccccc}
			\hline
			\multirow{2}{*}{Dataset} & \multicolumn{5}{c}{$\stab^q_n$ } \\
			 & RT & RuleFit & NodeHarvest & CA & SIRUS \\
			\hline
			Ozone & \textbf{1.0} & 0.11 & 0.92 & 0.24 & \textbf{0.99} \\
			Machine & 0.63 & 0.27 & \textbf{0.91} & 0.17 & \textbf{1.0} \\
			MPG & \textbf{1.0} & 0.14 & 0.87 & 0.25 & \textbf{1.0} \\
			Boston & 0.85 & 0.15 & 0.81 & 0.26 & \textbf{0.97} \\
			Student & \textbf{0.98} & 0.14 & \textbf{1.0} & 0.26 & \textbf{1.0} \\
			Abalone & \textbf{1.0} & 0.21 & 0.86 & 0.25 & \textbf{0.99} \\
			\hline
		\end{tabular}
		\vspace*{0.3cm}
		
		\begin{tabular}{lccccc}
			\hline
			\multirow{2}{*}{Dataset} & \multicolumn{5}{c}{$\simp_n$ } \\
			 & RT & RuleFit & NodeHarvest & CA & SIRUS \\
			\hline
			Ozone & 0.12 & 0.01 & 0.04 & \textbf{0.96} & 0.29 \\
			Machine & 0.14 & 0.02 & 0.04 &\textbf{ 0.9} & 0.25 \\
			MPG & 0.15 & 0.01 & 0.05 & \textbf{0.98} & 0.34 \\
			Boston & 0.26 & 0.01 & 0.07 & \textbf{1.0} & 0.52 \\
			Student & 0.37 & 0.05 & 0.25 & 0.91 & \textbf{0.97} \\
			Abalone & 0.58 & 0.02 & 0.13 & 0.66 & \textbf{1.0} \\
			\hline
		\end{tabular}
		\vspace*{0.3cm}
		
		\begin{tabular}{lccccc}
			\hline
			\multirow{2}{*}{Dataset} & \multicolumn{5}{c}{$\I$ } \\
			 & RT & RuleFit & NodeHarvest & CA & SIRUS \\
			\hline
			Ozone & 0.56 & 0.29 & 0.54 & \textbf{0.59} & \textbf{0.63} \\
			Machine & \textbf{0.52} & 0.41 & \textbf{0.56} & \textbf{0.55} & \textbf{0.57} \\
			MPG & \textbf{0.63} & 0.33 & 0.57 & 0.61 & \textbf{0.69} \\
			Boston & 0.57 & 0.3 & 0.52 & 0.5 & \textbf{0.69} \\
			Student & 0.47 & 0.12 & 0.49 & 0.43 & \textbf{0.74 }\\
			Abalone & 0.66 & 0.26 & 0.45 & 0.43 & \textbf{0.76} \\
			\hline
		\end{tabular}
 	
\end{table}


Another interesting result is obtained from the correlation matrix table \ref{tab:corr_reg}, which was calculated considering all results generated by the $10$-fold cross-validation for all datasets. It shows that the simplicity score is negatively correlated with the predictivity score, which illustrates the well-known predictivity / simplicity trade-off. Furthermore, the stability score seems to be uncorrelated with the predictivity score, but negatively correlated with the simplicity score, a result which is less expected.

\begin{table}
\centering
\caption{\label{tab:corr_reg} Correlation between the scores for the regressions experiments}
		\begin{tabular}{cccc}
			\hline
			 & $\pred_n$  & $\stab^q_n$ & $\simp_n$  \\
			\hline
			$\pred_n$ & $1$ & $-0.1$  & $-0.27$ \\
			$\stab^q_n$ & $-$ & $1$  &$-0.10$\\
			$\simp_n$ & $-$ & $-$  &$1$ \\
			\hline
		\end{tabular}	 
\end{table}

One may note that the distributions of the scores are very different. Indeed, the ranges for $q$-stability and simplicity are small relative to the predictivity scores. This may be explained by the fact that all algorithms are designed to be accurate, but not necessarily stable or simple. For example, SIRUS was thought to be stable, and according to the $q$-stability score it is with a score of about 1. On the other hand, stability was not considered for RuleFit and its $q$-stability score is always low. We can apply the same reasoning for the simplicity score.

\subsection{Results for classification}
\begin{figure}

\includegraphics[scale=0.45]{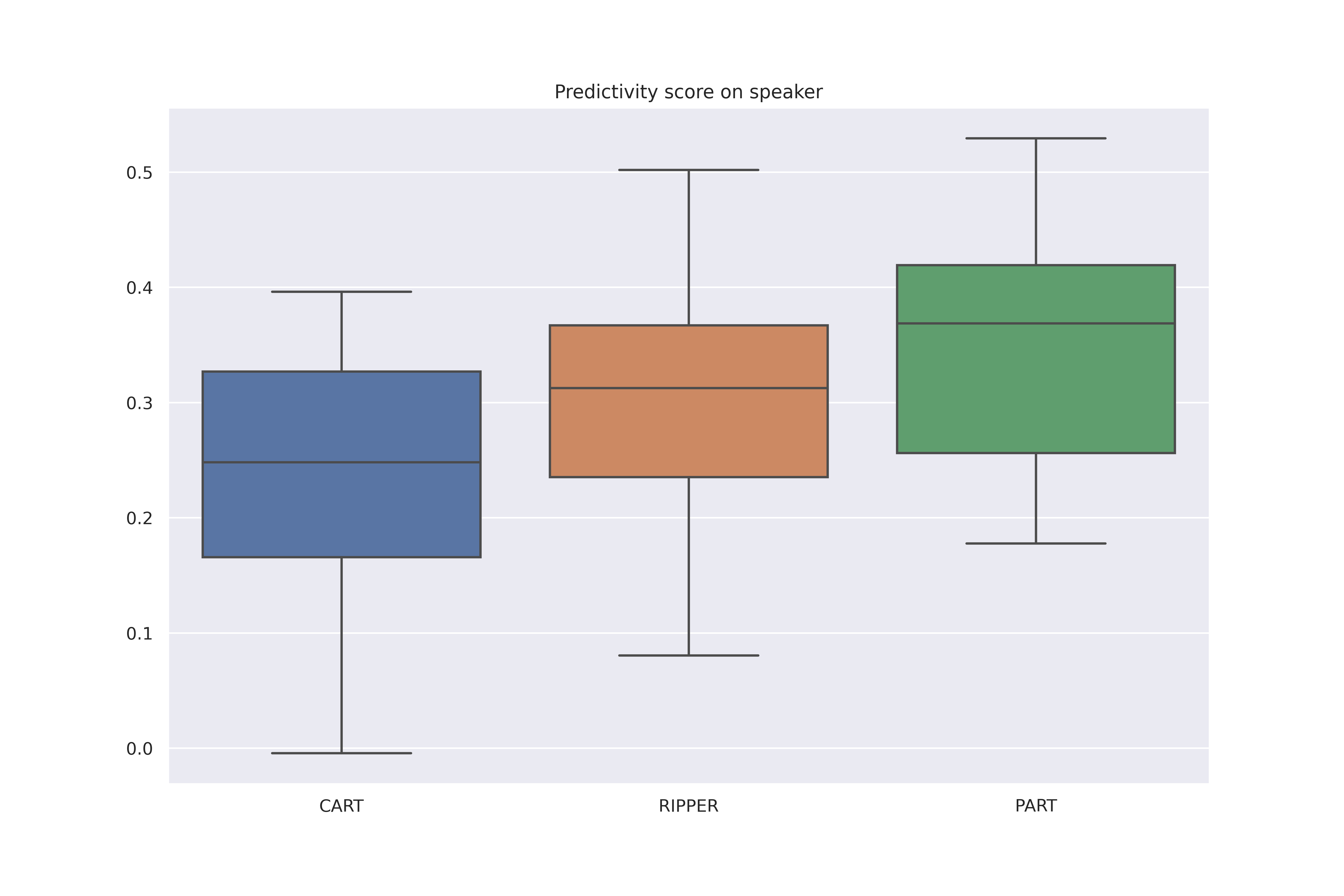}
\caption{\label{fig:pred2} Box-plot of the prediction scores for each algorithm for the dataset Speaker}
\end{figure}

\begin{figure}
\includegraphics[scale=0.45]{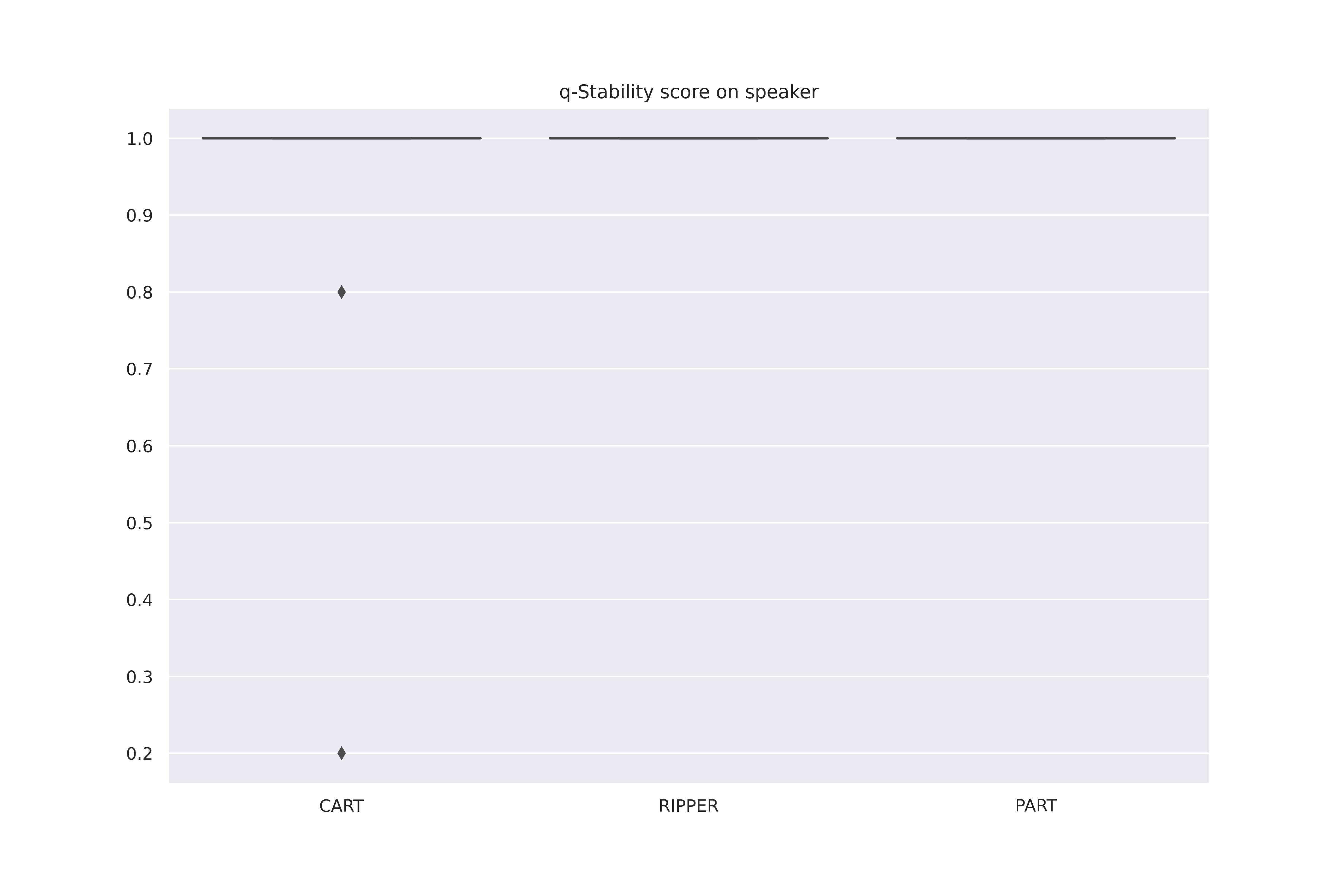}
\caption{\label{fig:stab2} Box-plot of the $q$-stability scores for each algorithm for the dataset Speaker}
\end{figure}

\begin{figure}

\includegraphics[scale=0.45]{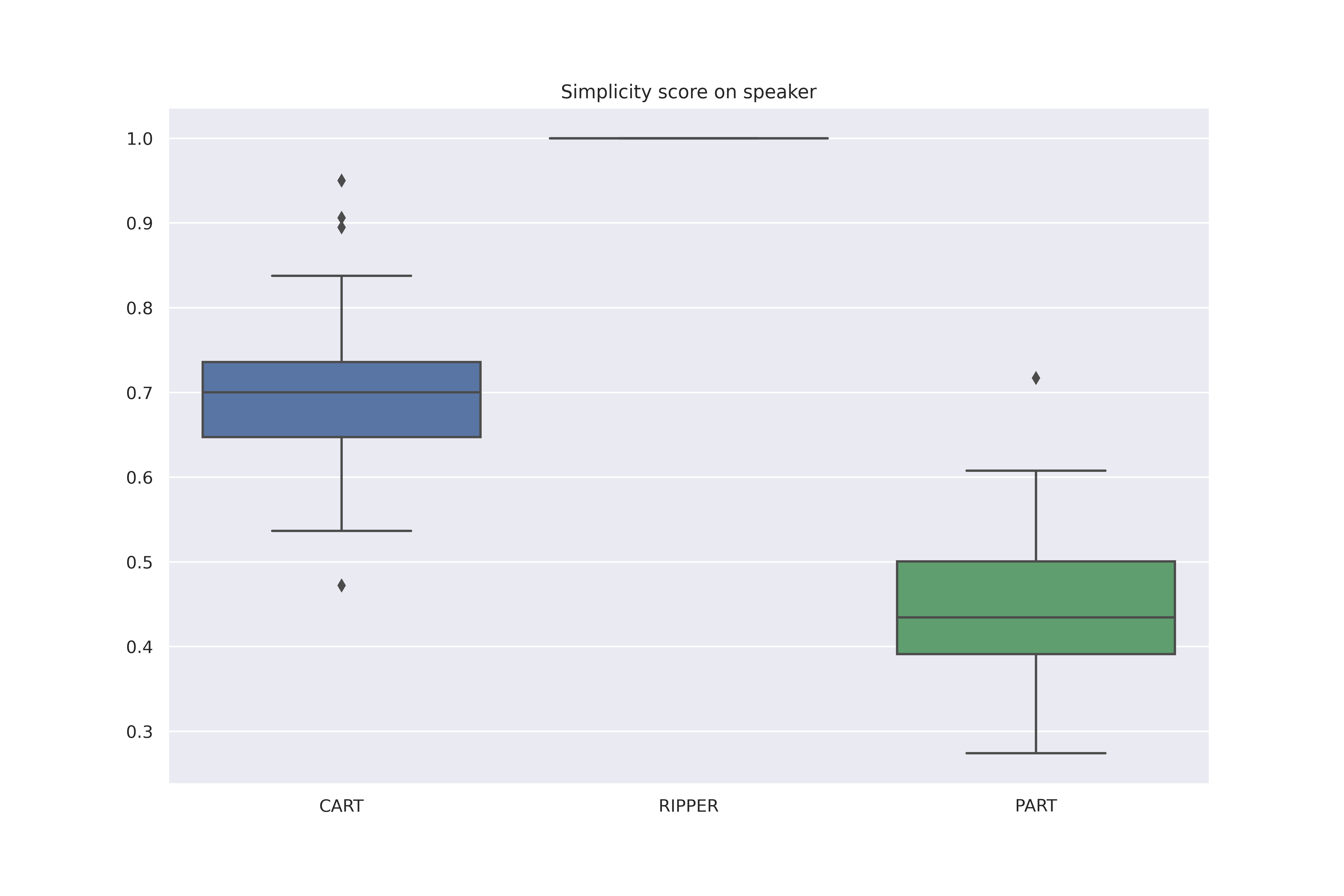}
\caption{\label{fig:simp2} Box-plot of the simplicity scores for each algorithm for the dataset Speaker}
\end{figure}

 	The averaged scores are summarized in Table \ref{tab:res_classif}. All selected algorithms have the same accuracy for all datasets. However, RIPPER and PART are both very stable algorithms, and RIPPER is the simplest of the three algorithms. Therefore, for these datasets and among these three algorithms, RIPPER is the algorithm that is most interpretable according to our measure \eqref{eq:interpretability}. Figures \ref{fig:pred2}, \ref{fig:stab2} and \ref{fig:simp2} are the box-plots of the predictivity scores, $q$-stability scores and simplicity scores, respectively, of each algorithms for the dataset Speaker.
 	
\begin{table}
\centering
\caption{\label{tab:res_classif} Average of predictivity score ($\pred_n$), stability score ($\stab_n^q$), simplicity score ($\simp_n$) and interpretability score ($\I$) over a $10$-fold cross-validation of commonly used interpretable algorithms for various public classification datasets. Best values are in bold, as well as values within 10\% of the maximum value for each dataset.}
		\begin{tabular}{lccc}
			\hline
			\multirow{2}{*}{Dataset} & \multicolumn{3}{c}{$\pred_n$ } \\
			 & CART & RIPPER & PART \\
			\hline
			Wine & \textbf{0.13} & \textbf{0.12} & 0.01 \\
			Covertype & 0.37 & \textbf{0.46} & \textbf{0.5} \\
			Speaker & 0.24 & 0.31 & \textbf{0.35} \\
			\hline
		\end{tabular}
	\vspace*{0.3cm}
	
		\begin{tabular}{lccc}
			\hline
			\multirow{2}{*}{Dataset} & \multicolumn{3}{c}{$\stab^q_n$ } \\
			 & CART & RIPPER & PART \\
			\hline
			Wine & \textbf{1.0} & \textbf{1.0} & \textbf{1.0} \\
			Covertype & \textbf{1.0} & \textbf{1.0} & \textbf{1.0} \\
			Speaker & 0.95 & \textbf{1.0} & \textbf{1.0} \\
			\hline
		\end{tabular}
	\vspace*{0.3cm}
		
	\begin{tabular}{lccc}
			\hline
			\multirow{2}{*}{Dataset} & \multicolumn{3}{c}{$\simp_n$ } \\
			 & CART & RIPPER & PART \\
			\hline
			Wine & \textbf{0.99} & 0.64 & 0.01 \\
			Covertype & \textbf{1.0} & 0.12 & 0.01 \\
			Speaker & 0.71 & \textbf{1.0} & 0.45\\
			\hline
	\end{tabular}
	\vspace*{0.3cm}
		
	\begin{tabular}{lccc}
			\hline
			\multirow{2}{*}{Dataset} & \multicolumn{3}{c}{$\I$ } \\
			 & CART & RIPPER & PART \\
			\hline
			Wine & \textbf{0.71} & 0.59 & 0.34 \\
			Covertype & \textbf{0.79} & 0.53 & 0.50 \\
			Speaker & 0.63 & \textbf{0.77} & 0.6 \\
			\hline
	\end{tabular}
 	
\end{table}

In contrast to the regression case, the correlation matrix table \ref{tab:corr_classif}, which was calculated considering all scores generated by the $10$-fold cross-validation for all datasets, shows that the scores do not seem to be correlated.

\begin{table}
\centering
\caption{\label{tab:corr_classif} Correlation between scores for the classification experiments}
		\begin{tabular}{cccc}
			\hline
			 & $\pred_n$  & $\stab^q_n$ & $\simp_n$  \\
			\hline
			$\pred_n$ & $1$ & $0.09$  & $ -0.04$ \\
			$\stab^q_n$ & $-$ & $1$  &$0.06$\\
			$\simp_n$ & $-$ & $-$  &$1$ \\
			\hline
		\end{tabular}
	 
\end{table}


These results should take into account that for the classification part we have tested fewer algorithms on fewer data sets than for the regression part. The accuracy of the models for these datasets was small. If the algorithms are not accurate enough, it may not be useful to look at the other scores. The algorithms appear to be very stable, which may be explained by the fact that they are not complex. Since it is not enough to have a good predictivity score, these algorithms must be tuned to be more complex.

\section{Conclusion and perspectives}
	In this paper we propose a score that may be used to compare the interpretability of tree-based algorithms and rule-based algorithms. This score is based on the triptych: predictivity \eqref{eq:pred}, stability \eqref{eq:stab}, and simplicity \eqref{eq:simp}, as proposed in \cite{Yu20,Benard21}. The proposed methodology seems to provide an easy way to rank the interpretability of a set of algorithms by being composed of three different scores that allow to integrate the main components of interpretability. It may be seen from our applications that the $q$-stability score and the simplicity score are quite stable regardless of the datasets. This observation is related to the properties of the algorithms; indeed, an algorithm designed for accuracy, stability or simplicity should maintain this property independent of the datasets.
	
	It is important to note that, according to the definition \ref{def:inter}, $100$ rules of length $1$ have the same interpretability index \eqref{sec:Inter} as a single rule of length $100$, which may be debatable. Furthermore, the stability score is purely syntactical and quite restrictive. If some features are duplicated, two rules can have two different syntactical conditions, but they are otherwise identical due to their activations. One possibility to relax the stability score could be to compare the rules on the basis of their activation sets (i.e. by searching for observations where the conditions are fulfilled simultaneously). Another issue is the selection of the weights in the interpretability formula \eqref{eq:interpretability}. For simplicity, we have used equal weights in this paper, but future work is needed on the optimal choice of these weights to match the specific goals of the analyst.

	As seen from the paper, the proposed interpretability score is meaningless unless it is used for the comparison of two or more algorithms. In future work, we intend to develop an interpretability score that can be computed for an algorithm regardless if other algorithms are considered or not. We also plan to adapt the measure of interpretability to other well-known ML algorithms and ML problems such as clustering or dimension reduction methods. To achieve this goal we will need to modify the definitions of the $q$-stability score and the simplicity score. Indeed, these two scores can be currently computed only for rule-based algorithms or tree-based algorithms (after a transformation of the generated tree into a set of rules). 
	
Another interesting extension would be the addition of a semantic analysis of the variables involved in the rules. In fact, NLP methods could be used to measure the distance between the target and these variables in a text corpus. This distance could be interpreted as the relevance of using such variables to describe the target.


\end{document}